\documentclass[sigconf]{acmart}

\copyrightyear{2021} 
\acmYear{2021} 
\setcopyright{acmcopyright}\acmConference[SIGIR '21]{Proceedings of the 44th International ACM SIGIR Conference on Research and Development in Information Retrieval}{July 11--15, 2021}{Virtual Event, Canada}
\acmBooktitle{Proceedings of the 44th International ACM SIGIR Conference on Research and Development in Information Retrieval (SIGIR '21), July 11--15, 2021, Virtual Event, Canada}
\acmPrice{15.00}
\acmDOI{10.1145/3404835.3462927}
\acmISBN{978-1-4503-8037-9/21/07}

\begin{document}
\fancyhead{} 
\title{Improving Video Retrieval by Adaptive Margin}


\author{Feng He, Qi Wang, Zhifan Feng, Wenbin Jiang, Yajuan Lü, Yong zhu, Xiao Tan }
\email{{hefeng07, wangqi31, fengzhifan, jiangwenbin, lvyajuan, zhuyong, tanxiao01}@baidu.com}
\affiliation{%
  \institution{Baidu Inc.}
  \city{Beijing}
  \country{China}
}


\begin{abstract}
Video retrieval is becoming increasingly important owing to the rapid emergence of videos on the Internet. The dominant paradigm for video retrieval learns video-text representations by pushing the distance between the similarity of positive pairs and that of negative pairs apart from a fixed margin. However, negative pairs used for training are sampled randomly, which indicates that the semantics between negative pairs may be related or even equivalent, while most methods still enforce dissimilar representations to decrease their similarity. This phenomenon leads to inaccurate supervision and poor performance in learning video-text representations. 

While most video retrieval methods overlook that phenomenon, we propose an adaptive margin changed with the distance between positive and negative pairs to solve the aforementioned issue. First, we design the calculation framework of the adaptive margin, including the method of distance measurement and the function between the distance and the margin. Then, we explore a novel implementation called "Cross-Modal Generalized Self-Distillation" (CMGSD), which can be built on the top of
most video retrieval models with few modifications. Notably, CMGSD adds few computational overheads at train time and adds no computational overhead at test time. Experimental results on three widely used datasets demonstrate that the proposed method can yield significantly better performance than the corresponding backbone model, and it outperforms state-of-the-art methods by a large margin.

\end{abstract}




\maketitle

\section{Introduction}
\label{section intro}
With the exponential growth of videos on the Internet, video retrieval is becoming increasingly important for search engines and multimedia data management\cite{wang2016comprehensive}. Traditional keyword-based video retrieval \cite{chang2015semantic,dalton2013zero,habibian2014composite} methods extract keywords from both video and text so that they can be compared directly. However, keyword-based approaches carry insufficient semantics due to the limited keywords. To address this problem, an increasing number of researchers focus on cross-modal video retrieval methods \cite{dong2019dual,mithun2018learning,chen2020fine,yu2018joint,yang2020tree,gabeur2020multi}, which use embeddings of video and text as their representations by leveraging the strong representation ability of deep neural networks.

\begin{figure}
\centering
\includegraphics[scale=0.2]{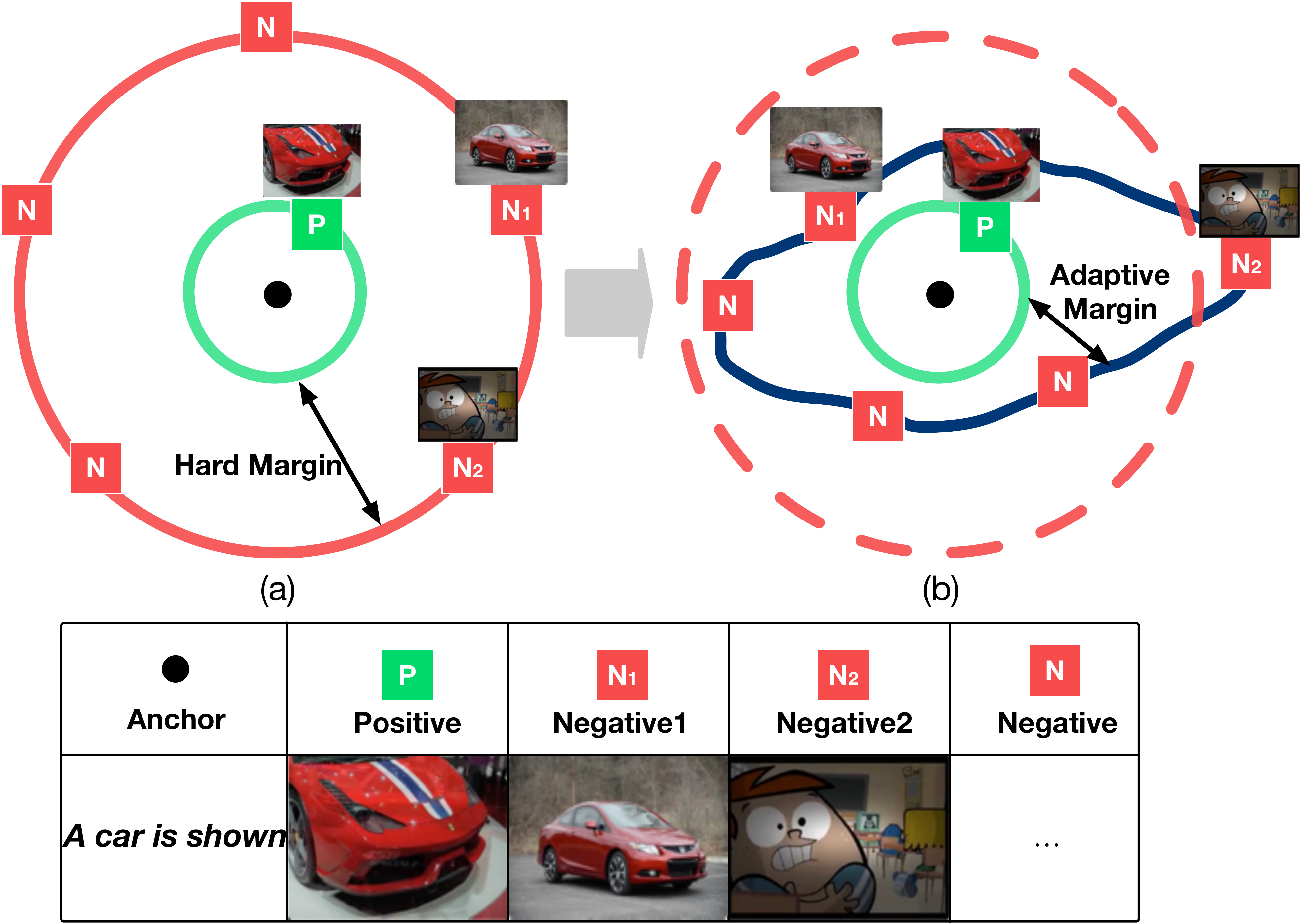}
\caption{Hard margin vs. Adaptive margin for video retrieval. (a) The red line is the hard margin, which clearly shows that all the negative samples need to be far away by a fixed distance. (b) The blue line is the adaptive margin changed with different negative samples. A dissimilar negative pair is assigned to a large margin, and a similar negative pair is assigned to a small margin.}
\label{fig:intro}
\end{figure}

The critical issue for cross-modal video retrieval is video-text representation learning for the comparison of video and text embeddings. For simplicity, the currently dominant approach can be summarized in two steps: 1) Encoding of video and text, which is to obtain video and text representations. 2) Video-text representation learning, wherein most methods utilize a triplet ranking loss as the objective function to provide supervision for learning video-text representations. Recently, efforts have been made to learn richer semantic representations of video and text. Dong \textit{et al.}\cite{dong2019dual} proposed a novel dual network that exploits multi-level encodings to obtain global, local, and temporal patterns in videos and sentences. Chen \textit{et al.}\cite{yang2020tree} proposed tree-augmented cross-modal encoding for video retrieval to obtain a richer representation of a complex query. Gabeur \textit{et al.}\cite{gabeur2020multi} introduced a transformer-based encoder architecture that effectively processes multiple modality features extracted from video to build effective video representations.

While most works focus on developing a stronger model for encoding video and text, the triplet ranking loss is employed to push away the representations of all negative pairs from the positive pair by a fixed margin called the "hard margin". That enforces similar representations for positive pairs and dissimilar representations for negative pairs. However, the negative pair is randomly sampled, which indicates that the training sample may provide inaccurate supervision for video-text representations, as shown in Figure \ref{fig:intro}(a).  It is natural that, compared to $N2$, a smaller margin is more suitable for $N1$ since the similarity between the anchor and $N1$ is much higher than the similarity between the anchor and $N2$. While $N1$ and the anchor share the same semantic content, most methods still push their distances away by the same margin, decreasing similarity between the anchor and $N1$. That makes $N1$ a tough sample to be optimized while imposing an extremely high penalty on it, ultimately leading to a wrong learning direction for video-text representations.

While most video retrieval methods overlook the above phenomenon, we propose a sample-specific adaptive margin changed with the different training samples to provide enhanced supervision for video-text representation learning. As shown in Figure \ref{fig:intro}(b), the adaptive margin can be regarded as a functional hard margin, whose input is the distance between positive and negative pairs, and the output is the margin that should be pulled apart. Specifically, we first assume the video and its corresponding text are semantically equivalent. Thus, the similarity distance between video and text across modalities can be approximated by the distance in the video domain and text domain, respectively. Based on this, we define several "\textit{supervision experts}" to explicitly measure the distance in two specific domains. Then, we propose a \textit{rescale function} to obtain the mapping relationship between the distance and the adaptive margin, making the adaptive margin fluctuate around the hard margin based on the distance between positive and negative pairs. From another perspective, the adaptive margin we proposed can also be regarded as a novel distillation framework that distills knowledge from single-modal domains to the cross-modal domain to provide enhanced supervision.

Moreover, we explore a novel implementation for our proposed framework, called \textit{Cross-Modal Generalized Self-Distillation} (CMGSD). Unlike the conventional self-distillation method, which directly feeds in predictions of the trained model as new target values for retraining, our approach uses common components of the video retrieval model to generate training targets indirectly. That also indicates that the proposed CMGSD can be built on the top of most video retrieval models. Notably, CMGSD adds few computational overheads at train time since it mainly utilizes several original outputs of the video retrieval model. Besides, CMGSD is only used at train time to provide enhanced supervision, which means it adds no computational overhead at test time.

To verify the effectiveness of our proposed method, we conducted extensive experiments on three popular datasets, MSRVTT\cite{xu2016msr}, ActivityNet\cite{huang2017densely}, and LSMDC\cite{rohrbach2015dataset}. Experimental results demonstrate that our methods can yield significantly better performance than the corresponding backbone model and outperform state-of-the-art methods by a large margin.

In brief, our contributions can be summarized as follows:
\begin{itemize}
\item We propose a sample-specific adaptive margin, which can be regarded as a functional hard margin. It is a framework that distills knowledge from two specific single-modal domains to provide enhanced supervision.

\item We propose the Cross-Modal Generalized Self-Distillation (CMGSD) method, which is a novel implementation for the adaptive margin. CMGSD is easy to be built on the top of most video retrieval models. Moreover, CMGSD adds few computational overheads at train time and adds no computational overhead at test time.

\item We conduct extensive experiments on three widely used datasets to demonstrate that our proposed method can significantly yield better performance than the corresponding backbone models and outperform state-of-the-art methods by a large margin.
\end{itemize}

\section{Related Work}
\paragraph{Video Retrieval}
Video retrieval has traditionally been a challenging task, and researchers have explored the idea of learning joint text-video representations, also known as common space learning, to improve retrieval performance. Most previous works focused on developing a stronger embedding method to obtain richer representations of video and text, respectively. Mithun \textit{et al.} \cite{mithun2018learning} employed multi-modal cues from different modalities in video to obtain a feature-richer video representation. Miech\textit{et al.}\cite{miech2018learning} proposed a Mixture-of-Embedding-Experts (MEE) model to learn text-video embeddings from heterogeneous data and handle missing input modalities. Liu \textit{et al.}\cite{liu2019use} proposed a collaborative experts framework for learning text-video embeddings by utilizing a gating mechanism, aggregating information from different features extracted from different modalities to obtain a richer representation of the video. Dong \textit{et al.}\cite{dong2019dual} proposed a multilevel dual encoder to encode sequential videos and texts. Chen \textit{et al.}\cite{chen2020fine} proposed a hierarchical graph reasoning (HGR) model that divided the representation of video and text into three levels for fine-grained video retrieval. Yang \textit{et al.}\cite{yang2020tree} developed a novel complex-query video retrieval framework to obtain a richer representation of a complex query, resulting in a performance boost. Gabeur \textit{et al.}\cite{gabeur2020multi} proposed a multi-modal transformer that allowed different modalities to consider the others, which helped achieve the state-of-the-art. 

However, the supervision provided by the triplet ranking loss, which is a common component of the video retrieval model\cite{mithun2018learning,miech2018learning,liu2019use,dong2019dual,chen2020fine,yang2020tree,gabeur2020multi}, sometimes leads to a wrong learning direction. Unfortunately, most methods did not pay attention to the supervision, which is critical for video-text representation learning. Therefore, in this work, we aim at providing enhanced supervision to improve the learning framework for video retrieval.

\paragraph{Triplet Loss with Adaptive Margin} 
Triplet loss\cite{weinberger2009distance} is a general framework for learning representations for different tasks. Conventional triplet loss utilizes a fixed margin to push positive and negative pairs apart, which indicates it treats different training samples equally. Recently, many scholars have improved the effect by changing the fixed margin to an adaptive margin\cite{cheng2016person,semedo2019cross,zakharov20173d,li2020symmetric,zhang2019learning,zhang2019learning,hu2018cvm}. Li \textit{et al.}\cite{li2020symmetric} proposed a novel Symmetic Metric Learning with an adaptive margin that was personalized for each user, improving the performance of the recommendation system. Zhang \textit{et al.}\cite{zhang2019learning} replaced the hard margin with a dynamically updated non-parametric adaptive margin that improved upon the state-of-the-art for both real-valued and binary descriptor learning. Hu \textit{et al.}\cite{hu2018cvm} introduced a new weighted adaptive margin ranking loss, speeding up the training convergence and improving image retrieval accuracy. While most adaptive margins were proposed in the uni-modal domain, Semedo \textit{et al.}\cite{semedo2019cross} proposed a pair-specific margin for image-text retrieval by category cluster and preservation. Note that this method was based on the semantic category of the image, which did not exist in video retrieval.

However, hardly any advanced adaptive margin mechanism has been proposed for video retrieval. In this work, we present an adaptive margin, providing enhanced supervision to improve video-text representations learning. The proposed adaptive margin is changed with different training pairs, which can also be regarded as a distillation framework for video retrieval. 

\paragraph{Self-Distillation} 
Knowledge distillation\cite{hinton2015distilling} is a framework to transfer knowledge from one to another. Self-distillation is a specific situation that transfers knowledge from the current training model to itself. Recently, self-distillation has been successfully applied to different research fields for improving the performance of different tasks. Hahn \textit{et al.}\cite{hahn2019self} unitized self-distillation to improve the performance of different NLP tasks. Zhang \textit{et al.}\cite{zhang2019your} proposed a self-distillation framework, improving the performance of image classification. Mobahi \textit{et al.} \cite{mobahi2020self} provided a theoretical analysis of self-distillation, which shown self-distillation may reduce over-fitting.

In this work, we apply self-distillation to video retrieval by exploring a novel implementation of the proposed adaptive margin. To our best knowledge, our proposed method is the first work that introduces self-distillation for providing enhanced supervision to video retrieval.

\section{Background}
\subsection{Problem Definition}
Given a data set $\mathcal D$ of video-text pairs $\{(v_i,t_i)\}$, where $v_i$ is a video and $t_i$ is a text that describes the content of the $v_i$, the task of video retrieval is to retrieve the most similar text(video) given video(text). Instead of optimizing retrieval results directly, most methods employ the triplet ranking loss as the objective function for learning joint video-text representations so that the video and text similarity can be compared directly. Let $r_v^i$ be the representation of $v_i$ and $r_t^i$ be a representation of $t_i$. Concretely, the triplet ranking loss for video retrieval is defined in the following form:
\begin{equation}
\begin{aligned}
\label{small_s}
    l_{v}^{h}(i,j;\alpha) = [s(r_v^j,r_t^i)-s(r_v^i,r_t^i)+\alpha]_+,
\end{aligned}
\end{equation}
\begin{equation}
\begin{aligned}
    l_{t}^{h}(i,j;\alpha) = [s(r_v^i,r_t^j)-s(r_v^i,r_t^i)+\alpha]_+,
\end{aligned}
\end{equation}
\begin{equation}
\begin{aligned}
    \mathcal L^h = &\frac{1}{B} \sum_{i=1}^B [\mathop{\max}\limits_{j}l_{v}^h(i,j;\alpha) + \mathop{\max}\limits_{j}l_{t}^h(i,j;\alpha)],
\end{aligned}
\end{equation}
where $s(\cdot,\cdot)$ represents the similarity measurement and $\alpha$ is the margin. $l_{v}^{h}(i,j;\alpha)$ and $l_{t}^{h}(i,j;\alpha)$ are the losses produced by two negative pairs $(r_v^j,r_t^i)$ and $(r_v^i,r_t^j)$, respectively, $\mathcal L$ is the overall triplet ranking loss. Given a positive pair $(v_i,t_i)$, we find its hardest negatives in a mini-batch and then push their distances from the positive pair further away than a fixed margin $\alpha$, called the "\textit{hard margin}" 

\subsection{The Architecture of Video Retrieval Models}
\subsubsection{Video Encoding}
For the video representation $r_v^i = \Psi(v_i) = \Psi''(\Psi'(v_i))$, the process of video encoding can be simply summarized in two steps:
\paragraph{Feature Extracting}
Most methods first use one or more pre-trained models trained on different labeled datasets and modalities to obtain effective multi-modal video features. Specifically, given $N$ pre-trained models $\{\psi'_n\}_{n=1}^N$, the process of extracting a video $v_i$ can be denoted as follows:
\begin{equation}
\begin{aligned}
    \Psi'(v_i) = [\psi'_1(v_i),...,\psi'_N(v_i)],
\end{aligned}
\end{equation}
where $\Psi^{'}(v_i)$ is the multi-modal feature of $v_i$. Notably, these pre-trained models are frozen during training.

\paragraph{Feature Aggregation}
After obtaining the multi-modal features of $v_i$, most methods utilize a video feature aggregator $\Psi''$ to obtain an aggregated video representation as follows:

\begin{equation}
\begin{aligned}
    r_v^i = \Psi''(\Psi'(v_i)).
\end{aligned}
\end{equation}
By aggregating the multi-modal feature of $v_i$, we obtain the aggregated video representation $r_v^i$.

\subsubsection{Text Encoding} Unlike the video representation obtained in two steps, most methods utilize a text encoder to directly extract the text representation. For simplicity, given a text $t_i$, the process is as follows:
\begin{equation}
\begin{aligned}
    r_t^i = \Phi(t_i),
\end{aligned}
\end{equation}
where $\Phi$ is the text encoder, $r_t^i$ is the representation of $t_i$.
\subsubsection{Similarity Measurement}
After obtaining the representation of video and text, the similarity measurement between them is defined as
\begin{equation}
\begin{aligned}
    s(r_v^i,r_t^i) = \frac{<r_v^i,r_t^i>}{||r_v^i|| ||r_t^i|| },
\end{aligned}
\end{equation}
where $<\cdot,\cdot>$ represents the dot product function and $||\cdot||$ is the $L_2$ norm function. During training, the similarity is used for calculating triplet ranking loss to learn text-video representations. During testing, the result of video retrieval is sorted by similarity.

\section{Method}
In the following sections, we first describe the calculation framework of the adaptive margin. The proposed cross-modal generalized self-distillation method will be elaborated later.

\subsection{Adaptive Margin for Video Retrieval}
\subsubsection{Adaptive Margin}
In this section, we first discuss how we calculate the similarity distance between positive and negative pairs in two specific single-modal domains. Then, we introduce the rescale function, which defines the mapping relationship between similarity distance and adaptive margin.
\label{section adaptive margin}
We claim that the margin used as the target for the similarity distance between positive and negative pairs should be a sample-specific adaptive margin. Specifically, based on the discussion in section \ref{section intro}, the adaptive margin should be proportional to the similarity distance. Thus, given a positive pair $(v_i,t_i)$ , two negative pairs $(v_i,t_j)$ and $(v_j,t_i)$, the above description can be formulated as
\begin{equation}
\label{pi1}
   \mathcal M_1(i,j) \propto [\Upsilon(v_i, t_i)-\Upsilon(v_j, t_i)],
\end{equation}
\begin{equation}
\label{pi2}
    \mathcal M_2(i,j) \propto [\Upsilon(v_i, t_i)-\Upsilon(v_i, t_j)],
\end{equation}
where $\mathcal M_1(i,j)$ and $\mathcal M_2(i,j)$ are adaptive margins defined for two negative pairs $(v_j,t_i)$ and $(v_i,t_j)$, and $\Upsilon$ is a model that can measure the similarity across modalities.

\begin{figure}
\centering
\includegraphics[scale=0.14]{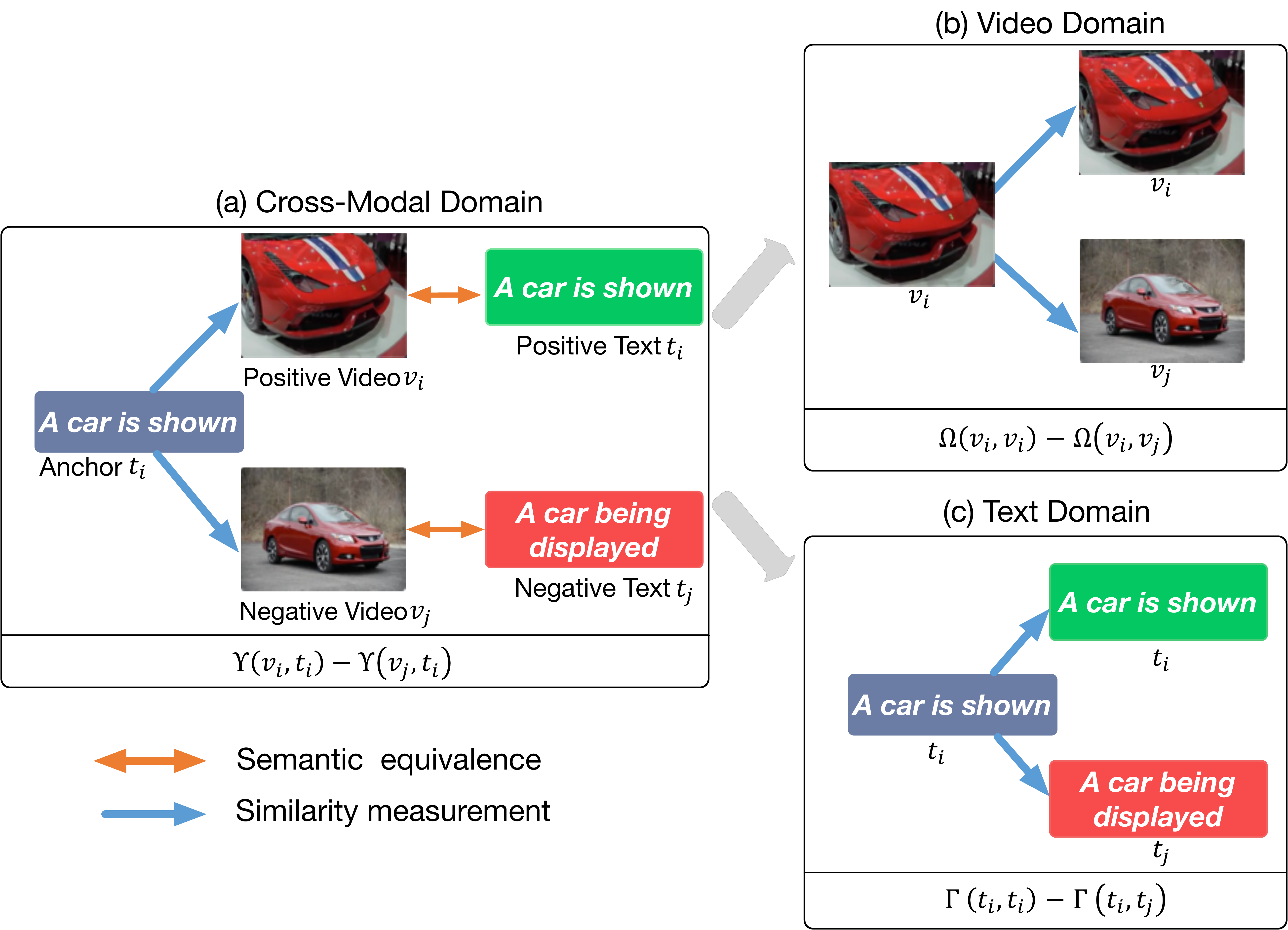}
\caption{Similarity measurement in different domains. (a) Similarity measurement in the cross-modal domain. (b) Similarity measurement in the video domain. (c) Similarity measurement in the text domain.}
\label{fig:single_model_domain}
\end{figure}

Instead of measuring similarity across modalities, we make a simple but effective assumption that can convert the similarity measurement in the cross-modal domain to the similarity measurement in two specific single-modal domains. Our idea is based on the fact that the video $v_i$ and its corresponding text $t_i$ can be regarded as semantic equivalent. Thus, the similarity distance between positive and negative pairs across modalities can be reflected by their similarity distance in the video and text domains, respectively, as shown in Figure \ref{fig:single_model_domain}.

To measure the similarity in the text domain and video domain, we introduce several powerful models with rich representations for text and video called "supervision experts". For notational simplicity, let $\Gamma$ be the text supervision expert used for measuring text similarity and $\Omega$ be the video supervision expert for measuring video similarity. Given a positive pair $(v_i,t_i)$, two negative pairs $(v_j,t_i)$ and $(v_i,t_j)$, we have the following form:
\begin{equation}
\begin{aligned}
\label{text_domain}
   [\Upsilon(v_i, t_i)-\Upsilon(v_j, t_i)] &\propto [\Omega(v_i, v_i)-\Omega(v_i, v_j)],
\end{aligned}
\end{equation}
\begin{equation}
\begin{aligned}
\label{video_domain}
   [\Upsilon(v_i, t_i)-\Upsilon(v_j, t_i)] &\propto [\Gamma(t_i, t_i)-\Gamma(t_i, t_j)],
\end{aligned}
\end{equation}
\begin{equation}
\begin{aligned}
\label{text_domain2}
    [\Upsilon(v_i, t_i)-\Upsilon(v_i, t_j)] &\propto [\Omega(v_i, v_i)-\Omega(v_i, v_j)],
\end{aligned}
\end{equation}
\begin{equation}
\begin{aligned}
\label{video_domain2}
    [\Upsilon(v_i, t_i)-\Upsilon(v_i, t_j)] &\propto [\Gamma(t_i, t_i)-\Gamma(t_i, t_j)].
\end{aligned}
\end{equation}
Notably, unlike cross-modal similarity measurement, there are many available tools\cite{devlin2019bert,xie2018rethinking,he2016deep,reimers2019sentence} for obtaining the similarity in the single-modal domain. 

Combining Eq. \ref{text_domain} to \ref{video_domain2} and Eq. \ref{pi1} to \ref{pi2}, we obtain the following formulation:
\begin{equation}
\begin{aligned}
    \mathcal M_1(i,j) &\propto \Omega(v_i, v_i)-\Omega(v_i, v_j)\\
             &\propto \Gamma(t_i, t_i)-\Gamma(t_i, t_j),
\end{aligned}
\end{equation}
\begin{equation}
\begin{aligned}
    \mathcal M_2(i,j) &\propto \Omega(v_i, v_i)-\Omega(v_i, v_j) \\
             &\propto \Gamma(t_i, t_i)-\Gamma(t_i, t_j),
\end{aligned}
\end{equation}
Clearly, $\mathcal M_1(i,j)$ and $\mathcal M_2(i,j)$ can be reflected by the same similarity. Thus, we can simplify them as follows:
\begin{equation}
\begin{aligned}
\mathcal M(i,j) &\propto \Omega(v_i, v_i)-\Omega(v_i, v_j) \\
                &\propto \Gamma(t_i, t_i)-\Gamma(t_i, t_j),
\end{aligned}
\end{equation}
where $\mathcal M(i,j)$ is the adaptive margin for both $(v_j,t_i)$ and $(v_i,t_j)$. It is obvious that the proposed adaptive margin can be obtained by comparing the similarity in the video and text domains, respectively.
\begin{figure}
\centering
\includegraphics[scale=0.35]{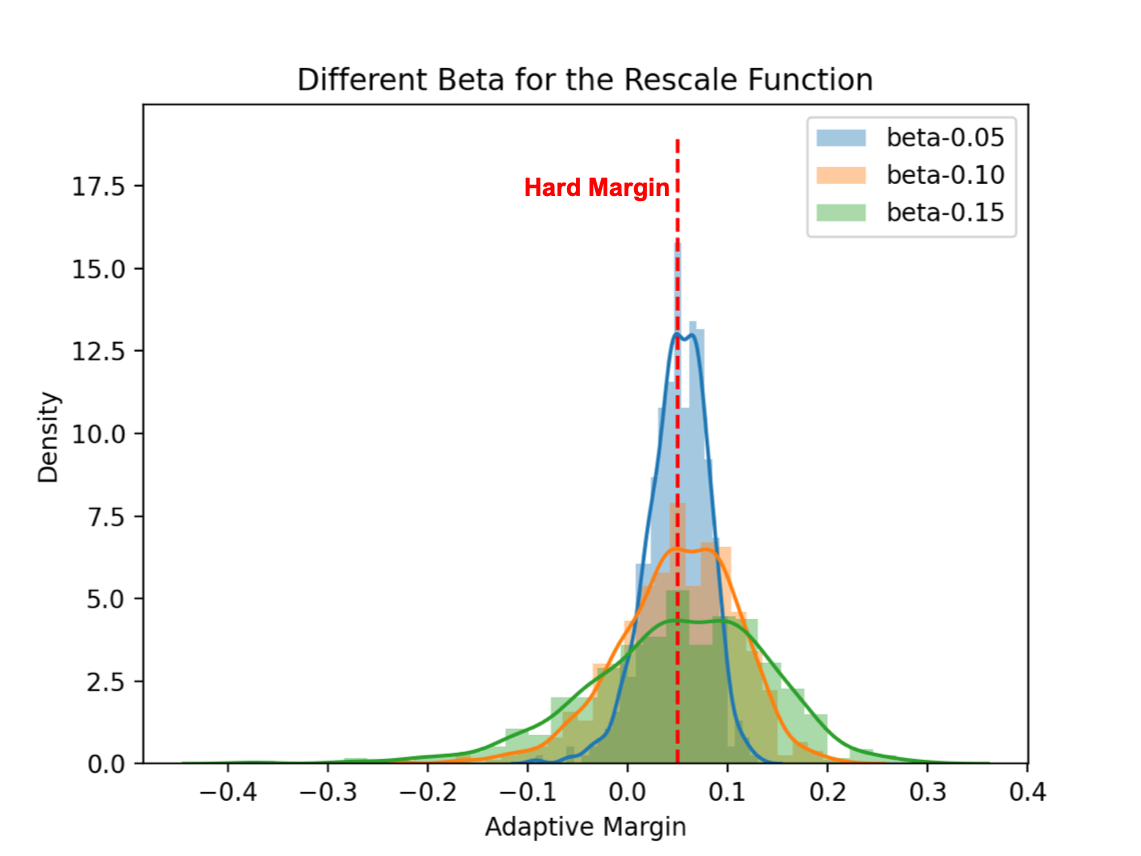}
\caption{Probability distributions of adaptive margins with different $\beta$ when hard margin is 0.05. Specifically, setting $\beta$ to 0.05 means 90\% adaptive margins in $[0,0.1]$. }
\label{fig:scale_func}
\end{figure}

\subsubsection{Rescale Function} 
\label{section Rescale Function}
\begin{figure*}[ht]
\centering
\includegraphics[width=1\textwidth]{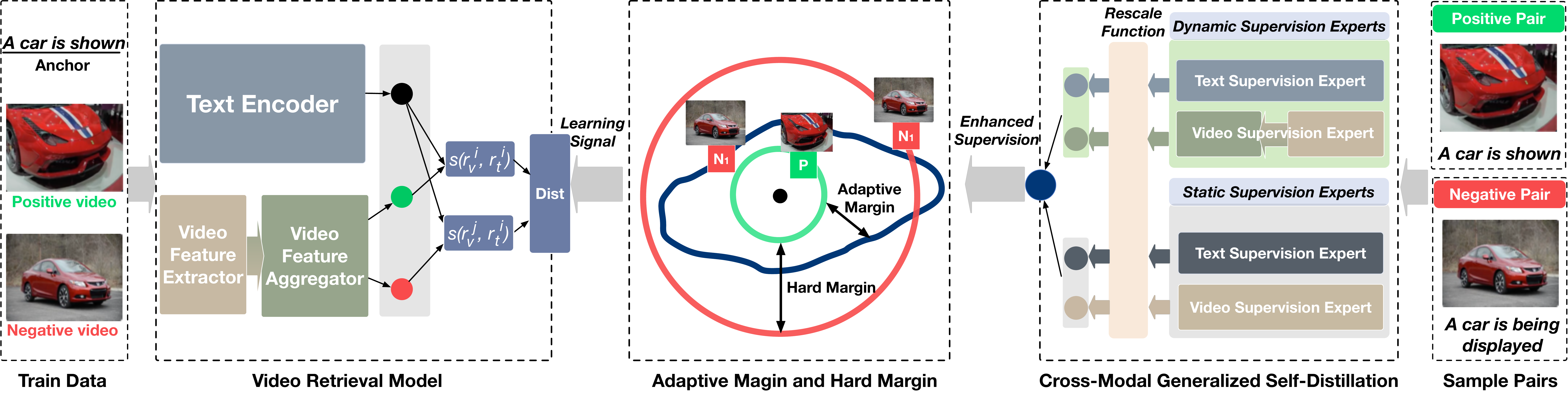}
\caption{Our cross-modal generalized self-distillation method. The left side of the figure shows the common architecture for video retrieval. The right side of the figure is our proposed method, which provides the adaptive margin for video-text representation learning. The center side of the figure represents the supervision signal provided by the adaptive margin and hard margin together. Notably, dynamic supervision experts and static supervision experts utilize common components of the video retrieval model, which are represented by the same color in the figure. 
}
\label{fig:model_arch}
\end{figure*}
We explicitly measure the similarity distance between positive and negative pairs in two single-modal domains by introducing several supervision experts. Further, we assume that the similarity distance between different positive and negative pairs is Gaussian. For notational simplicity, let $\Omega(i,j)$ and $\Gamma(i,j)$ be short of for $\Gamma(t_i, t_i)-\Gamma(t_i, t_j)$ and $\Omega(v_i, v_i)-\Omega(v_i, v_j)$, respectively. The similarity distance is a Gaussian means:
\begin{equation}
\begin{aligned}
\Omega(i,j) \sim \mathcal N (E[\Omega(i,j)],Var[\Omega(i,j)]),
\end{aligned}
\end{equation}
\begin{equation}
\begin{aligned}
\Gamma(i,j) \sim \mathcal N (E[\Gamma(i,j)],Var[\Gamma(i,j)]),
\end{aligned}
\end{equation}
where $\mathcal N(a,b)$ is the normal distribution with a mean of $a$ and variance of $b$, $E[\cdot]$ represents the expectation function, and $Var[\cdot]$ represents the variance function. Note that $E$ and $Var$ can be achieved by the estimations of mean and variance in a mini-batch\cite{ioffe2015batch}. Based on the above assumption, $\Omega(i,j)$ and $\Gamma(i,j)$ fluctuate around $E[\Omega(i,j)]$ and $E[\Gamma(i,j)]$, respectively, and their variances control fluctuation ranges. 

Following the above discussion, we introduce a simple yet effective rescale function, which defines the mapping relationship between similarity and adaptive margin as follows:
\begin{equation}
\begin{aligned}
\mathcal M_v(i,j)=\mathcal F(\Omega(i,j);\mu_v, \beta_v) = \frac{\Omega(i,j)-E[\Omega(i,j)]}{\sqrt{Var[\Omega(i,j)]/\mathcal U(\beta_v) }}+\mu_v,
\end{aligned}
\end{equation}
\begin{equation}
\begin{aligned}
 \mathcal M_t(i,j)= \mathcal F(\Gamma(i,j);\mu_t, \beta_t) = \frac{\Gamma(i,j)-E[\Gamma(i,j)]}{\sqrt{Var[\Gamma(i,j)]/\mathcal U(\beta_t)}}+\mu_t,
\end{aligned}
\end{equation}
where $\mathcal M$ is the adaptive margin, $\mathcal F$ is our proposed rescale function, $\mu$ and $\beta$ are parameters that control the fluctuation range of the adaptive margin relative to the hard margin, and $\mathcal U$ is a function achieved by a binary search that transforms $\beta$ into the corresponding variance. Note that we define two adaptive margins in both the video domain and text domain, which are expressed in the formula as subscripts $v$ and $t$. Clearly, $\mathcal F$ rescales the original distributions of similarity distance to the new normal distribution, which can be briefly summarized as
\begin{equation}
\begin{aligned}
\mathcal M_v(i,j) \sim \mathcal N(\mu_v, U(\beta_v)),
\end{aligned}
\end{equation}
\begin{equation}
\begin{aligned}
\mathcal M_t(i,j) \sim \mathcal N(\mu_t, U(\beta_t)).
\end{aligned}
\end{equation}
We set $\mu_{t}$ and $\mu_{v}$ to the hard margin to ensure that adaptive margins fluctuate around that. Instead of setting the variance of adaptive margin directly, which is not an intuitive way to control the fluctuation range of adaptive margins, we utilize a rescale factor $\beta$ to set the half size of the 90\% confidence interval of the normal distribution, as shown in Figure \ref{fig:scale_func}.

From another perspective, our proposed adaptive margin can also be regarded as a novel distance-based cross-modal distillation framework. The supervision expert used for measuring similarity distance can be regarded as the teacher model that distills knowledge from two specific single-modal domains to the cross-modal domain. Moreover, the rescale function can be compared with the temperature parameter $T$ in conventional knowledge distillation\cite{hinton2015distilling}. When our $\beta$ is close to zero, the adaptive margin and hard margin are almost equal. Meanwhile,  the adaptive margin does not provide additional valuable information compared to the hard margin. When $\beta$ gradually increases, the similarity information contained in the adaptive margin is gradually released. At this stage, the information extracted by the supervision expert has a greater impact on the video retrieval model.

\subsection{Cross-Modal Generalized Self-Distillation}
We explore a novel implementation called "Cross-Modal Generalized Self-Distillation" (CMGSD) under our proposed framework. CMGSD utilizes common components of the video retrieval model to define supervision experts, distilling knowledge from the current training model. We introduces two valuable components of the video retrieval model used in CMGSD  in Section \ref{section dse} and Section \ref{section sse}. Then, we describe the objective function of CMGSD in Section \ref{section objection} in detail. The architecture of CMGSD is presented in Figure \ref{fig:model_arch}. 

\subsubsection{Dynamic Supervision Experts}
\label{section dse}
The text and video encoders are common components of the current training model. Since the premise for the alignment between video and text is rich representations for both, the text encoder and video encoder obtain that ability implicitly with the training process. Thus, we can use them to measure the similarity distance between positive and negative pairs. Because they change with the training process, we call them "Dynamic Supervision Experts"(DSEs). Specifically, DSEs can be formulated as
\begin{equation}
\begin{aligned}
    \Gamma_{d}(t_i,t_i)- \Gamma_{d}(t_i,t_j) = 1 -\frac{<\Phi(t_i),\Phi(t_j)>}{||\Phi(t_i)|| ||\Phi(t_j)||} ,   
\end{aligned}
\end{equation}

\begin{equation}
\begin{aligned}
    \Omega_{d}(v_i,v_i)- \Omega_{d}(v_i,v_j) = 1-\frac{<\Psi(v_i),\Psi(v_j)>}{||\Psi(v_i)|| ||\Psi(v_j)||},
\end{aligned}
\end{equation}
where $\Phi(t_i)$ and $\Phi(t_j)$ are representations of $t_i$ and $t_j$ produced by the current training text encoder $\Phi$ , $\Psi(v_i)$ and $\Psi(v_j)$ are representations of $v_i$ and $v_j$ produced by the current training video encoder $\Psi$, $\Gamma_{d}$ and $\Omega_{d}$ are the text supervision expert and video supervision expert, respectively. Note that the results of $\Gamma_{d}(t_i,t_i)$ and $\Omega_{d}(t_i,t_i)$ are 1.

Notably, DSEs cannot provide an accurate measure of similarity at the beginning of the training process since it does not contain relevant information at that stage. However, the effect of DSEs continues to improve with the training process.
\subsubsection{Static Supervision Experts.}
\label{section sse}
Recall that most methods utilize one or more pre-trained models trained on different labeled datasets and tasks to embed video information from multiple modalities. That indicates each pre-trained model can provide the semantic representation from different perspective. Thus, we define static supervision experts (SSEs) by those pre-trained models. For notational simplicity, we assume only one pre-trained model $\psi'_n(v_i)$ is used to measure similarity, which is shown as
\begin{equation}
\begin{aligned}
 \Omega_{s}(v_i,v_i)-\Omega_{s}(v_i,v_j)= 1-\frac{<Pooling(\psi'_n(v_i)),Pooling(\psi'_n(v_j))>}{||Pooling(\psi'_n(v_i))|| ||Pooling(\psi'_n(v_j))||},
\end{aligned}
\end{equation}
where $\psi'_n(v_i)$ and $\psi'_n(v_j)$ are representations of $v_i$ and $v_j$ produced by the pre-trained model $\psi'_n$, and $\Omega_{s}$ is the video supervision expert. Because the feature extracted by the pre-training model is often frame-level, we use pooling to encode them to a fixed-length vector. Specifically, mean pooling is employed in our experiment. However, unlike DSEs, which provide adaptive margins in both text and video domains, SSEs can only provide adaptive margin in the video domain since no pre-trained model is used for text feature extraction. To overcome this shortcoming, we employ Sentence-Bert\cite{reimers2019sentence}, which can be regarded as a tool-kit for measuring similarity between texts, for providing supervision from the text domain as follows:

\begin{equation}
\begin{aligned}
    \Gamma_{s}(t_i,t_i)- \Gamma_{s}(t_i,t_j) = 1 -\frac{<Bert(t_i),Bert(t_j)>}{||Bert(t_i)|| ||Bert(t_j)||},
\end{aligned}
\end{equation}
where $Bert(t_i)$ is the representation of $t_i$ produced by Sentence-Bert, and $\Gamma_{s}$ is the text supervision expert defined by Sentence-Bert. Although Sentence-Bert does not belong to the common architecture of the video retrieval model, it is a convenient open-source toolkit that can be easily integrated into existing video retrieval models. 

Notably, the components of SSEs are trained out of the current domain, which indicates that the similarity measurement provided by SSEs may not be as accurate as DSEs in the later stages of the training process. However, SSEs can provide adaptive margins during the entire training process compared to DSEs.

\subsubsection{Leveraging Self-Knowledge to Training}
\label{section objection}
 We first define the triplet ranking loss by adaptive margins, which are shown as
\begin{equation}
\begin{aligned}
    l_{v}^{s}(i,j;\Omega,\Gamma) = &l_{v}^h(i,j;\mathcal F(\Omega(i,j;\alpha, \beta)))+ \\
    &l_{v}^h(i,j;\mathcal F(\Gamma(i,j;\alpha, \beta)), 
\end{aligned}
\end{equation}
\begin{equation}
\begin{aligned}
    l_{t}^{s}(i,j;\Omega,\Gamma)  =  &l_{t}^h(i,j;\mathcal F(\Omega(i,j;\alpha, \beta))+, \\
    &l_{t}^h(i,j;\mathcal F(\Gamma(i,j;\alpha, \beta)), 
\end{aligned}
\end{equation}
where $l_{v}^{s}$ and $l_{t}^{s}$ is the soft-margin-based triplet ranking loss with two negative pairs $(v_j,t_i)$ and $(v_i,t_j)$, respectively. 

To provide enhanced supervision for video-text representations, we regard $l_{v}^{s}$ and $l_{t}^{s}$ as distillation losses adding them to the standard triplet ranking loss, as shown below:
\begin{equation}
\begin{aligned} 
   \mathcal  L_{v}^{full}  = \frac{1}{B} \sum_{i=1}^B \mathop{\max}\limits_{j}[l_{v}^{h}(i,j;m) +\lambda&*l_{v}^{s}(i,j;\Omega_d,\Gamma_d) \\ +(1-\lambda)&*l_{v}^{s}(i,j;\Omega_s,\Gamma_s)],
\end{aligned}
\end{equation}
\begin{equation}
\begin{aligned}
   \mathcal L_{t}^{full}  = \frac{1}{B} \sum_{i=1}^B \mathop{\max}\limits_{t}[l_{t}^{h}(i,j;m) +\lambda&*l_{t}^{s}(i,j;\Omega_d,\Gamma_d) \\ +(1-\lambda)&*l_{t}^{s}(i,j;\Omega_s,\Gamma_s)],
\end{aligned}
\end{equation}
\begin{equation}
\begin{aligned}
    \mathcal L^{full}  = \mathcal L_{t}^{full} + \mathcal L_{v}^{full},
\end{aligned}
\end{equation}
where $\mathcal L_{v}^{full}$ and $\mathcal L_{t}^{full}$ are triplet losses for negative pair $(v_j,t_i)$ and negative pair $(v_i,t_j)$ after adding the distillation loss, $\mathcal L^{full}$ is the overall objective function, and $\lambda$ is a hyper-parameter used for balancing losses produced by SSEs and DSEs. In the initial stage of training, DSEs do not contain relevant information for calculating adaptive margins. Thus, $\lambda$ will be a small value, indicating that the SSEs mainly provide enhanced supervision at that time. With the training process, the effect of DSEs gradually exceed that of SSEs, so $\lambda$ gradually becomes larger. See more implementation details in Section \ref{section details}.

\section{EXPERIMENTAL SETUP}

\subsection{Research Questions}
We conducted experiments to evaluate the proposed method on three datasets. Mainly, we aim to answer the following research questions:
\begin{itemize}
\item Q1: Can our proposed CMGSD outperform state-of-the-art video retrieval methods?
\item Q2: What are the impacts of different components on the overall performance of our approach?
\item Q3: Can backbone models with our proposed method be better supervised than those without?
\item Q4: Can our proposed method be a useful tool-kit for most video retrieval models?

\end{itemize}

\subsection{Datasets}
We verified our proposed method on three datasets: MSRVTT\cite{xu2016msr}, ActivityNet\cite{krishna2017dense}, and LSMDC\cite{rohrbach2015dataset}.

\paragraph{MSRVTT\cite{xu2016msr}:} It is a popular dataset for video retrieval, which contains 10000 videos. Each video is 10 to 30 seconds long and is annotated with 20 descriptions. For this dataset, there are three different dataset partitions. The first one is the official partition from \cite{xu2016msr} with 6513 clips for training, 497 clips for validation, and the remaining 2990 clips for testing. The second one is from \cite{miech2018learning} with 9000 clips for training and 1000 clips for testing. The last one is the partitions from \cite{yu2018joint}, which uses 9000 clips for training and 1000 clips for testing. To increase the persuasiveness of the experiment, we conducted experiments on all data partitions. 

\paragraph{ActivityNet\cite{krishna2017dense}:} It is an increasingly popular dataset consisting of densely annotated temporal segments of 20000 YouTube videos. Each video is an average of 2 minutes long, and there are 72000 video-text pairs in this dataset. We follow the approach proposed in \cite{zhang2018cross}, which concatenates all the descriptions of a video to a paragraph. We trained the model with 10009 videos and evaluated our model on the "val1" split, which contains 4917 videos.

\paragraph{LSMDC\cite{rohrbach2015dataset}:} It contains 118,081 short video clips(~ 45 s) extracted from 202 movies. Unlike other datasets, each video in LSMDC only has one description, either extracted from the movie script or the transcribed audio description. The test set consisted of 1000 videos that were not present in the training set.

\subsection{Evaluation Metrics}
We evaluate our method with the following standard metrics: recall at rank K(R@K), median rank(MdR), and the sum of recalls(Rsum) which is the sum of all R@K. The higher R@K and Rsum mean the better performance of the model. Conversely, the lower MdR means better performance of the model. For each experiment, we report the mean and standard deviation over experiments with five random seeds.

\subsection{Implementation Details}
\label{section details}
\subsubsection{Backbone Model}
We used the multi-modal transformer (MMT) \cite{gabeur2020multi} as the backbone model to report our main results on three datasets. MMT is a transformer-based framework that jointly encodes the different modalities in the video, which allows each of them to attend to the others. Our CMGSD is built on the top of MMT and provides enhanced supervision to it. We retained all of the setting and training details of the backbone model. Note that we trained MMT with the hard negative mining, which was different from the original MMT. We trained MMT without hard negative mining for one epoch to warm up since we found that directly training with hard negative mining leads to optimization failures. After warming up tirck, the hard negative mining strategy improves the performance of MMT, which can be observed in Table \ref{as_modality}.

\subsubsection{Dynamic Supervision Experts}
The text dynamic supervision expert is the text encoder of MMT\cite{gabeur2020multi} itself. Specifically, we used the "CLS" token as the semantic representation to measure the similarity distance between positive and negative pairs. For the video dynamic supervision expert, we utilized pooling heads of appearance and motion from MMT to measure similarity distance in the video domain.
\subsubsection{Static Supervision Experts}
The text static supervision expert was achieved by Sentence-Bert\cite{reimers2019sentence}. Specifically, we used the "roberta-large-nli-stsb-mean-tokens" to calculate the similarity between texts. For the video static supervision expert, we utilized SENet-154\cite{hu2018squeeze} trained for classification on ImageNet\cite{deng2009imagenet} and S3D\cite{xie2018rethinking} trained on the Kinetics action recognition to measure the similarity between videos. Note that these two models are also used for video feature extraction in MMT, which indicates that they can be regarded as components of MMT.
\subsubsection{Hyperparameters}
 For three datasets, we set the same hyperparameters. Precisely, to control the fluctuation range of adaptive margin,  we set $\beta$ to 0.04. For warm-up DSEs, we first set $\lambda$ to 0 at epochs 1 to 20. At the 20$th$ epoch, we set $\lambda$ to 0.1, and it becomes 1.0 at the 50$th$ epoch through exponential growth.

\section{EXPERIMENTAL RESULTS}
\begin{table*}[ht]
  \caption{Retrieval performance on the MSRVTT dataset. We quantified the performance of our method on three different dataset partitions. "1K-A" denoted dataset partition used in \cite{yu2018joint}. "1K-B" denoted dataset partition used in \cite{miech2018learning}. "Full" denoted dataset partition used in \cite{xu2016msr}. Top scores were highlighted.}
  \label{msrvtt}
  \footnotesize
  \renewcommand\arraystretch{0.85}
  \begin{tabular}{c|c|cccc|cccc|c}
    \toprule
    \multicolumn{2}{c}{} & \multicolumn{4}{c}{text-to-video} & \multicolumn{4}{c}{video-to-text} \\
    Method  & Split& $R@1\uparrow$ & $R@5\uparrow$ & $R@10\uparrow$ & $MdR\downarrow$ & $R@1\uparrow$ & $R@5\uparrow$ & $R@10\uparrow$ & $MdR \downarrow$ &$Rsum\uparrow$\\
    \midrule
    JSFusion \cite{yu2018joint}  &1k-A &10.2 &31.2 &43.2 &13.0 &- &-&-&- &-\\
    HT\cite{miech2019howto100m} &1k-A &12.1 &35.0 &48.0 &12.0 &-&-&-&-\\
    JPoSE \cite{wray2019fine} &1k-A &14.3 &38.1 &53.0 &9.0  &16.4 &41.3 &54.4 & 8.7  &217.5\\
    MME\cite{miech2018learning} &1k-A &16.8 &41.0 &54.4 &9.0  &- &- &- &- &-\\
    TCE\cite{yang2020tree} &1k-A &17.1 &39.9 &53.7 &9.0  &-&-&-&-&-\\
    CE\cite{liu2019use} &1k-A &$20.9_{\pm 1.2}$ &$48.8_{\pm 0.6}$ &$62.4_{\pm 0.8}$ &$6.0_{\pm 0}$  &$20.6_{\pm 0.6}$ &$50.3_{\pm 0.5}$ &$64.0_{\pm 0.2}$ &$5.3_{\pm 0.6}$ &$267.0$\\
    MMT\cite{gabeur2020multi}&1k-A &$24.6_{\pm 0.4}$ &$54.0_{\pm 0.2}$ &$67.1_{\pm 0.5}$ &$4.0_{\pm 0}$  &$24.4_{\pm 0.5}$ &$56.0_{\pm 0.9}$ &$67.8_{\pm 0.3}$ &$4.0_{\pm 0.0}$  &$293.9$\\
    Ours &1k-A &$\textbf{26.1}_{\pm 0.2}$ &$\textbf{56.8}_{\pm 0.3}$ &$\textbf{69.7}_{\pm 0.4}$ &$\textbf{4.0}_{\pm 0.0}$ &$\textbf{27.2}_{\pm 0.6}$ &$\textbf{58.0}_{\pm 0.7}$ &$\textbf{69.5}_{\pm 0.7}$ &$\textbf{3.9}_{\pm 0.1}$  &$\textbf{307.2}$\\
    \hline
    MME\cite{miech2018learning} &1k-B &13.6 &37.9 &51.0 &10.0  &- &- &- &- \\
    MEE-COCO\cite{miech2018learning}&1k-B &14.2 &39.2 &53.8 &9.0 &-&-&-&- &-\\
    JPose\cite{wray2019fine} &1k-B &14.3 &38.1 &53.0 &9.0 &16.4 &41.3 &54.4 &8.7 &217.5\\
    TCE\cite{yang2020tree}&1k-B &17.1 &39.9 &53.7 &9.0 &- &- &- &- &-\\
    CE\cite{liu2019use} &1k-B &$18.2_{\pm 0.7}$ &$46.0_{\pm 0.4}$ &$60.7_{\pm 0.2}$ &$7.0_{\pm 0.0}$ &$18.0_{\pm 0.8}$ &$46.0_{\pm 0.5}$ &$60.3_{\pm 0.5}$ &$6.5_{\pm 0.5}$  &$249.2$\\
    MMT\cite{gabeur2020multi}&1k-B &$20.3_{\pm 0.5}$ &$49.1_{\pm 0.4}$ &$63.9_{\pm 0.5}$ &$6.0_{\pm 0.0}$ &$21.1_{\pm 0.4}$ &$49.4_{\pm 0.4}$ &$63.2_{\pm 0.4}$ &$6.0_{\pm 0.0}$ &$267.0$\\
    Ours &1k-B &$\textbf{22.7}_{\pm 0.7}$ &$\textbf{52.6}_{\pm 1.7}$ &$\textbf{66.1}_{\pm 1.4}$ &$\textbf{4.6}_{\pm 0.3}$ &$\textbf{23.1}_{\pm 1.1}$ &$\textbf{53.5}_{\pm 1.3}$ &$\textbf{64.3}_{\pm 0.90}$ &$\textbf{4.6}_{\pm 0.3}$  &$\textbf{282.4}$\\
    \hline
    VSE\cite{kiros2014unifying} &Full &5.0 &16.4 &24.6 &47.0 &7.7 &20.3 &31.2 &28.0 & 105.2\\
    VSE++\cite{faghri2017vse++} &Full &5.7 &17.1 &24.8 &65.0  &10.2 &25.4 &35.1 &25.0  & 118.3\\
    W2VV \cite{dong2018predicting} &Full &6.1 &18.7 &27.5 &45.0  &11.8 &28.9 &39.1 &21.0  & 132.1\\
    Dual\cite{dong2019dual} &Full &7.7 &22.0 &31.8 &32.0  &13.0 &30.8 &43.3 &15.0 & 148.6\\
    TCE\cite{yang2020tree} &Full &7.7 &22.5 &32.1 &30.0 &- &- &- &-  &-\\
    HGR \cite{chen2020fine} &Full &9.2 &26.2 &36.5 &24.0 &15.0 &36.7 &48.8 &11.0   & 172.4\\
    CE\cite{liu2019use} &Full &$10.0_{\pm 0.1}$ &$29.0_{\pm 0.3}$ &$41.2_{\pm 0.2}$ &$16.0_{\pm 0}$   &$15.6_{\pm 0.3}$ &$40.9_{\pm 1.4}$ &$55.2_{\pm 1.0}$ &$8.3_{\pm 0.6}$  &$191.9$\\
    Ours &Full &$\textbf{11.3}_{\pm 0.0}$ &$\textbf{32.0}_{\pm 0.0}$ &$\textbf{44.1}_{\pm 0.0}$ &$\textbf{14.2}_{\pm 0.2}$  &$\textbf{17.2}_{\pm 0.8}$ &$\textbf{43.6}_{\pm 0.3}$ &$\textbf{57.2}_{\pm 0.2}$ &$\textbf{7.6}_{\pm 0.3}$ &$\textbf{205.4}$\\
    \bottomrule
  \end{tabular}
\end{table*}

\begin{table*}[ht]
  \caption{Retrieval performance on the ActivityNet dataset.}
  \label{activitynet}
  \footnotesize
  \renewcommand\arraystretch{0.85}
  \begin{tabular}{c|cccc|cccc|c}
    \toprule
    \multicolumn{1}{c}{} & \multicolumn{4}{c}{text-to-video} & \multicolumn{4}{c}{video-to-text} \\
    Method  & $R@1\uparrow$ & $R@5\uparrow$ & $R@50\uparrow$ & $MdR\downarrow$ & $R@1\uparrow$ & $R@5\uparrow$ & $R@50\uparrow$ & $MdR \downarrow$ &$Rsum\uparrow$\\
    \midrule
    FSE \cite{zhang2018cross}&$18.2_{\pm 0.2}$ &$44.8_{\pm 0.4}$ &$89.1_{\pm 0.3}$ &$7.0_{\pm 0.0}$ &$16.7_{\pm 0.8}$ &$43.1_{\pm 1.1}$ &$88.4_{\pm 0.3}$ &$7.0_{\pm 0.0}$ &$300.3$\\
    CE \cite{liu2019use}&$18.2_{\pm 0.3}$ &$47.7_{\pm 0.6}$ &$91.4_{\pm 0.4}$ &$6.0_{\pm 0.0}$  &$17.7_{\pm 0.6}$ &$46.6_{\pm 0.7}$ &$90.9_{\pm 0.2}$ &$6.0_{\pm 0.0}$  &$312.5$\\
    HSE \cite{zhang2018cross} &$20.5$ &$49.3$ &- &- &$18.7$ &$48.1$ &- &- &-\\
    MMT \cite{gabeur2020multi}&$22.7_{\pm 0.2}$ &$54.2_{\pm 1.0}$ &$93.2_{\pm 0.4}$ &$5.0_{\pm 0.0}$  &$22.9_{\pm 0.8}$ &$54.8_{\pm 0.4}$ &$93.1_{\pm 0.2}$ &$4.3_{\pm 0.5}$ &$340.7$\\
    Ours&$\textbf{24.2}_{\pm 0.1}$ &$\textbf{56.3}_{\pm 0.3}$ &$\textbf{94.0}_{\pm 0.00}$ &$\textbf{4.0}_{\pm 0.0}$ &$\textbf{24.6}_{\pm 0.2}$ &$\textbf{56.8}_{\pm 0.5}$ &$\textbf{93.8}_{\pm 0.0}$ &$\textbf{4.0}_{\pm 0.0}$  &$\textbf{349.7}$\\
    \bottomrule
  \end{tabular}
\end{table*}

\begin{table*}[ht]
  \caption{Retrieval performance on the LSMDC dataset.}
  \label{lsmdc}
  \footnotesize
  \renewcommand\arraystretch{0.85}
  \begin{tabular}{c|cccc|cccc|c}
    \toprule
    \multicolumn{1}{c}{} & \multicolumn{4}{c}{text-to-video} & \multicolumn{4}{c}{video-to-text} \\
    Method  & $R@1\uparrow$ & $R@5\uparrow$ & $R@10\uparrow$ & $MdR\downarrow$ & $R@1\uparrow$ & $R@5\uparrow$ & $R@10\uparrow$ & $MdR \downarrow$ &$Rsum\uparrow$\\
    \midrule
    CT-SAN\cite{yu2017end} &5.1 &16.3 &25.2 &46 &- &- &- &- &-\\
    CCA\cite{klein2015associating} &7.5 &21.7 &31.0 &33  &- &- &- &- &-\\
    JSFusion\cite{yu2018joint} &9.1 &21.2 &34.1 &36  &- &- &- &- &-\\
    MEE\cite{miech2018learning} &9.3 &25.1 &33.4 &27  &- &- &- &- &-\\
    MEE-COCO\cite{miech2018learning} &10.1 &25.6 &34.6 &27  &- &- &- &- &-\\
    TCE\cite{yang2020tree} &10.6 &25.8 &35.1 &29  &- &- &- &- &-\\
    CE\cite{liu2019use} &$11.2_{\pm 0.4}$ &$26.9_{\pm 1.1}$ &$34.8_{\pm 2.0}$ &$25.3_{\pm 3.1}$  &- &-  &-   &-\\
    MMT\cite{gabeur2020multi}&$13.2_{\pm 0.4}$ &$29.2_{\pm 0.8}$ &$38.8_{\pm 0.9}$ &$21.0_{\pm 1.4}$ &$12.1_{\pm 0.1}$ &$29.3_{\pm 1.1}$ &$37.9_{\pm 1.1}$ &$22.5_{\pm 0.4}$ &$160.5$\\
    Ours&$\textbf{14.0}_{\pm 0.5}$ &$\textbf{31.1}_{\pm 0.0}$ &$\textbf{41.0}_{\pm 0.7}$ &$\textbf{18.3}_{\pm 0.5}$ &$\textbf{13.0}_{\pm 0.3}$ &$\textbf{30.6}_{\pm 0.5}$ &$\textbf{39.6}_{\pm 0.2}$ &$\textbf{21.2}_{\pm 2.1}$ &$\textbf{169.3}$\\
    \bottomrule
  \end{tabular}
\end{table*}
\subsection{Comparison to State-of-The-Arts}
In this subsection, we answer Q1: We evaluated our proposed method by comparing it against state-of-the-art video retrieval methods on three datasets. 

Table \ref{msrvtt} summarizes the performance comparison results between our proposed CMGSD with state-of-the-art methods. Our model surpasses state-of-the-art methods on three dataset partitions by a large margin for both text-to-video and video-to-text retrieval. In particular, our method exceeded text-to-video retrieval of current state-of-the-art methods by 1.5, 2.4, and 1.3 on R@1 for three partitions, respectively. Moreover, our method improved R@1 by 2.8, 2.0, and 1.6 for video-to-text retrieval.

Table \ref{activitynet} shows retrieval results on the ActivityNet dataset. Overall, we observed similar patterns in terms of the overall performance in Table \ref{msrvtt}. Specifically, the overall retrieval quality reflected by the Rsum metric was boosted by a large margin (+9.0). Since videos in the ActivityNet dataset have a longer length and texts are much more complicated than MSRVTT, the results in Table \ref{activitynet} demonstrated the robustness of our approach on the different types of video and text.

Table \ref{lsmdc} shows retrieval results on the LSMDC dataset. With the adaptive margin that provided enhanced supervision for video-text representations, our method also outperformed previous state-of-the-art methods by a margin on all evaluation metrics.

\begin{table*}[ht]
  \caption{Ablation studies of supervision experts from different modalities. We quantified the contributions of
the text supervision experts(TSEs) and video supervision experts(VSEs), respectively.}
  \label{as_modality}
  \footnotesize
  \renewcommand\arraystretch{0.85}
  \begin{tabular}{c|ccc|cccc|cccc|c}
    \toprule
    \multicolumn{4}{c}{} & \multicolumn{4}{c}{text-to-video} & \multicolumn{4}{c}{video-to-text} \\
   Row&  TSEs & VSEs & Hard Mining & $R@1\uparrow$ & $R@5\uparrow$ & $R@10\uparrow$ & $MdR\downarrow$ & $R@1\uparrow$ & $R@5\uparrow$ & $R@10\uparrow$ & $MdR \downarrow$ &$Rsum\uparrow$\\
    \midrule
    0  &$\times$  & $\times$  & $\times$  &$24.6_{\pm 0.4}$ &$54.0_{\pm 0.2}$ &$67.1_{\pm 0.5}$ &$4.0_{\pm 0.0}$ &$24.4_{\pm 0.5}$ &$56.0_{\pm 0.9}$ &$67.8_{\pm 0.3}$ &$3.9_{\pm 0.1}$ &$293.9$\\
    1  &$\times$  & $\times$  & $\checkmark$  &$25.9_{\pm 0.3}$ &$56.3_{\pm 0.3}$ &$68.8_{\pm 1.0}$ &$4.0_{\pm 0.0}$ &$26.4_{\pm 0.9}$ &$56.6_{\pm 0.3}$ &$68.4_{\pm 1.0}$ &$4.0_{\pm 0.0}$ &$302.4$\\
    2  &$\checkmark$ &$\times$  & $\checkmark$ &$26.0_{\pm 0.5}$ &$56.5_{\pm 0.4}$ &$69.2_{\pm 0.4}$ &$4.0_{\pm 0.0}$  &$26.7_{\pm 0.8}$ &$56.7_{\pm 0.3}$ &$69.5_{\pm 1.9}$ &$4.0_{\pm 0.0}$  &$304.6$\\
    3 &$\times$  &$\checkmark$  & $\checkmark$ &$26.0_{\pm 0.4}$ &$\textbf{56.9}_{\pm 1.4}$ &$69.7_{\pm 0.3}$ &$4.0_{\pm 0.3}$ &$27.0_{\pm 0.7}$ &$57.4_{\pm 0.7}$ &$69.2_{\pm 0.5}$ &$4.0_{\pm 0.0}$  &$306.2$\\
    4 &$\checkmark$ &$\checkmark$  & $\checkmark$ &$\textbf{26.1}_{\pm 0.2}$ &$56.8_{\pm 0.3}$ &$\textbf{69.7}_{\pm 0.4}$ &$\textbf{4.0}_{\pm 0.0}$  &$\textbf{27.2}_{\pm 0.6}$ &$\textbf{58.0}_{\pm 0.7}$ &$\textbf{69.5}_{\pm 0.7}$ &$\textbf{3.9}_{\pm 0.1}$  &$\textbf{307.2}$\\
    \bottomrule
  \end{tabular}
\end{table*}

\begin{table*}[ht]
  \caption{Ablation studies to investigate contributions of static and dynamic supervision experts. SSEs was short for static supervision experts, and DSEs was short for dynamic supervision experts.}
  \label{as_distillation}
  \footnotesize
  \renewcommand\arraystretch{0.85}
  \begin{tabular}{c|cc|cccc|cccc|c}
    \toprule
    \multicolumn{3}{c}{} & \multicolumn{4}{c}{text-to-video} & \multicolumn{4}{c}{video-to-text} \\
    Row& SSEs & DSEs & $R@1\uparrow$ & $R@5\uparrow$ & $R@10\uparrow$ & $MdR\downarrow$ & $R@1\uparrow$ & $R@5\uparrow$ & $R@10\uparrow$ & $MdR \downarrow$ &$Rsum\uparrow$\\
    \midrule
    1  &$\times$  & $\times$  &$25.9_{\pm 0.3}$ &$56.3_{\pm 0.3}$ &$68.8_{\pm 1.0}$ &$4.0_{\pm 0.0}$  &$26.4_{\pm 0.9}$ &$56.6_{\pm 0.3}$ &$68.4_{\pm 1.0}$ &$4.0_{\pm 0.0}$ &$302.4$\\
    2 &$\checkmark$  & $\times$  &$25.8_{\pm 0.5}$ &$56.0_{\pm 1.4}$ &$69.6_{\pm 0.3}$ &$4.0_{\pm 0.0}$  &$26.6_{\pm 1.2}$ &$56.9_{\pm 0.8}$ &$69.0_{\pm 0.5}$ &$4.0_{\pm 0.0}$  &$303.9$\\
    3 &$\times$  & $\checkmark$ &$25.9_{\pm 0.3}$ &$\textbf{57.1}_{\pm 0.9}$ &$69.3_{\pm 0.2}$ &$4.0_{\pm 0.0}$  &$27.1_{\pm 0.9}$ &$57.1_{\pm 0.5}$ &$\textbf{69.7}_{\pm 0.1}$ &$4.0_{\pm 0.0}$  &$306.2$\\
    4 &$\checkmark$  & $\checkmark$ &$\textbf{26.1}_{\pm 0.2}$ &$56.8_{\pm 0.3}$ &$\textbf{69.7}_{\pm 0.4}$ &$\textbf{4.0}_{\pm 0.0}$ &$\textbf{27.2}_{\pm 0.6}$ &$\textbf{58.0}_{\pm 0.7}$ &$69.5_{\pm 0.7}$ &$\textbf{3.9}_{\pm 0.1}$ &$\textbf{307.2}$\\
    \bottomrule
  \end{tabular}
\end{table*}
\begin{table*}[ht]
  \caption{Other methods with CMGSD.}
  \footnotesize
  \renewcommand\arraystretch{0.85}
  \begin{tabular}{c|c|cccc|cccc|c}
    \toprule
    \multicolumn{2}{c}{} & \multicolumn{4}{c}{text-to-video} & \multicolumn{4}{c}{video-to-text} \\
    Method &CMGSD & $R@1\uparrow$ & $R@5\uparrow$ & $R@10\uparrow$ & $MdR\downarrow$ & $R@1\uparrow$ & $R@5\uparrow$ & $R@10\uparrow$ & $MdR \downarrow$ &$Rsum\uparrow$\\
    \midrule
    Dual\cite{dong2019dual} &$\times$ &$7.7$ &$22.0$ &$31.8$ &$32.0$ &$13.0$ &$30.8$ &$43.3$ &$15.0$ &$148.6$\\
    Dual\cite{dong2019dual} &$\checkmark$ &$\textbf{8.1}$ &$\textbf{23.4}$ &$\textbf{33.2}$ &$\textbf{29.0}$ &$\textbf{13.2}$ &$\textbf{32.1}$ &$\textbf{44.6}$ &$\textbf{14.0}$ &$\textbf{154.3}$\\
    \hline
    CE\cite{liu2019use} &$\times$ &$10.0$ &$29.0$ &$41.2$ &$16.0$ &$15.6$ &$40.9$ &$55.2$ &$8.3$ &$191.8$\\
    CE\cite{liu2019use} &$\checkmark$ &\textbf{11.8} &\textbf{32.5} &\textbf{44.2} &\textbf{14.0} &\textbf{17.7} &\textbf{44.9} &\textbf{59.2} &\textbf{7.0} &\textbf{210.3}\\
    \bottomrule
  \end{tabular}
  \label{Other method}
\end{table*}
\subsection{Ablation Studies}
In this section, we answer Q2: To investigate contributions of different components in our proposed method, we conducted several ablation studies on MSRVTT with the dataset partition from \cite{yu2018joint}. Note that, because our method uses hard mining, which is not used in the original MMT, the effect was somewhat different from \cite{gabeur2020multi} after removing all components.

\subsubsection{Supervision Experts from Different Modalities} Table \ref{as_modality} shows that enhanced supervision from both the text domain and the video domain is beneficial for video retrieval. With supervision provided by text supervision experts, the model gained a 2.2 improvement on Rsum comparison to the model without that. Similarly, the model achieved an improvement of 3.8 on Rsum with enhanced supervision provided by video supervision experts. Table \ref{as_modality} also demonstrated that the enhanced supervision provided by video supervision experts and text supervision experts were complementary, so the effect of the model further improved after supervision signals of text and video were merged.

\subsubsection{Dynamic and Static Supervision Experts} We evaluated the effort of dynamic and static supervision experts in Table \ref{as_distillation}, which shows that knowledge from both static supervision experts and dynamic supervision experts improved the performance of video retrieval. Also, we found that the supervision from in-domain(dynamic supervision experts) knowledge had a more positive impact on retrieval results than the supervision from out-domain knowledge(static supervision experts). Meanwhile, the effect was cumulative. After adding two kinds of supervision experts, the performance was maximized.

\subsection{Qualitative Results of the Adaptive Margin} 
\begin{figure}[ht]
\centering
\includegraphics[scale=0.25]{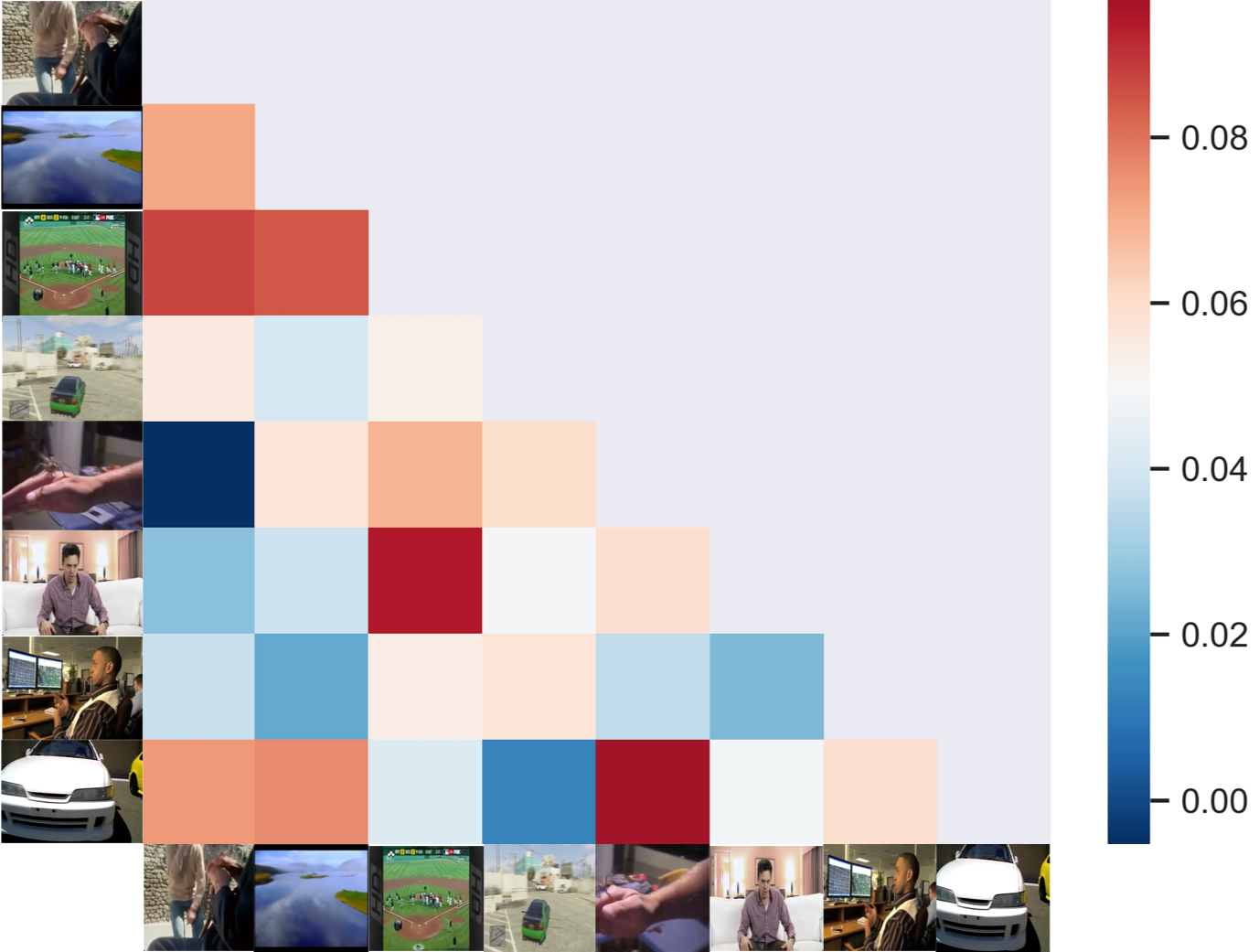}
\caption{Examples of the adaptive margin provided by the video dynamic supervision expert.}
\label{soft_margin1}
\end{figure}

\begin{figure}[ht]
\centering
\includegraphics[scale=0.22]{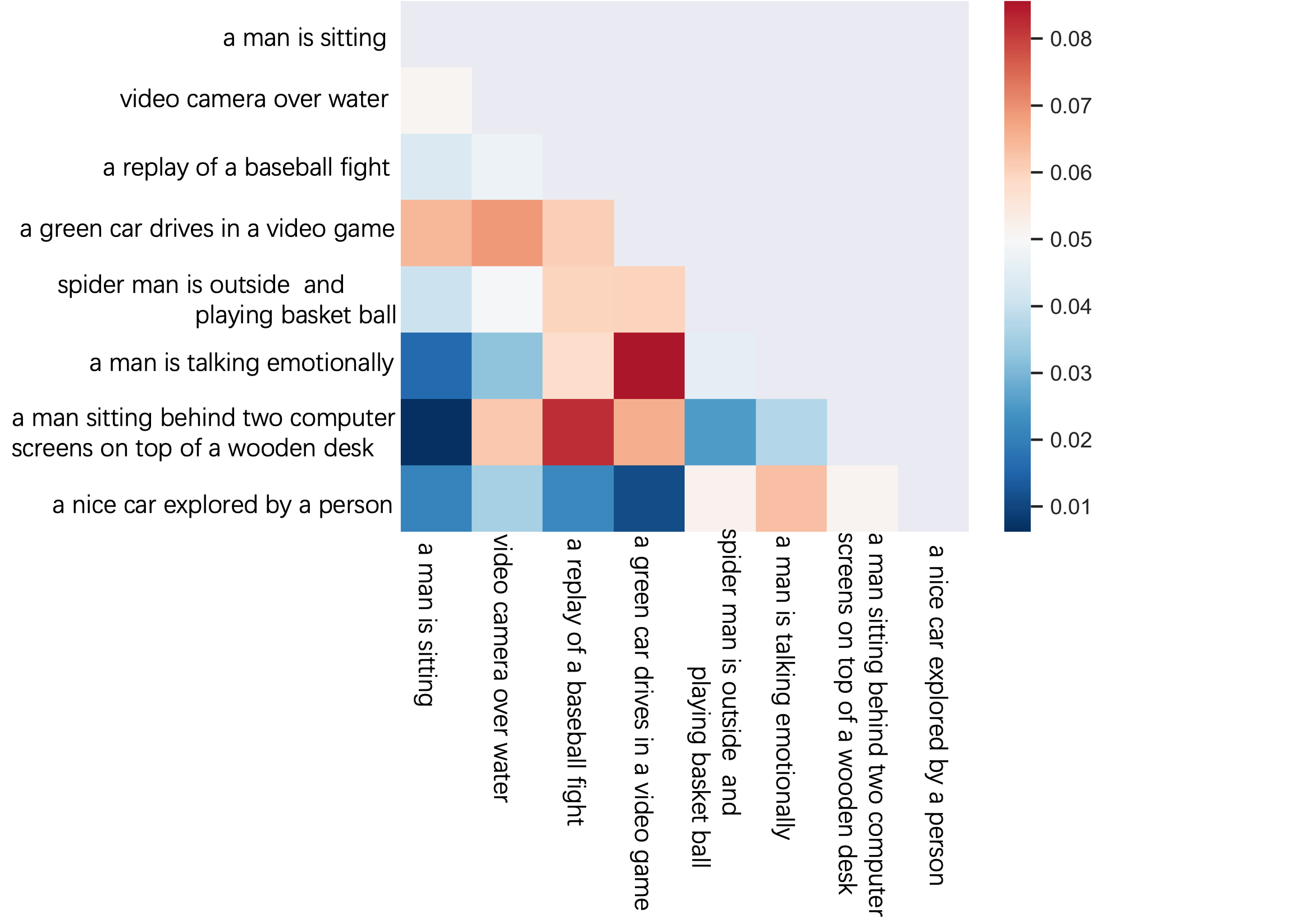}
\caption{Examples of the adaptive margin provided by the text dynamic  supervision expert.}
\label{soft_margin2}
\end{figure}

In this section, we answer Q3: To better understand the adaptive margin we proposed, we visualized the adaptive margin produced by DSEs in Figure \ref{soft_margin1} and Figure \ref{soft_margin2}. We observed that both the text dynamic supervision expert and the video dynamic supervision expert could provide more accurate supervision for video-text representation learning than the hard margin. Specifically, a dissimilar negative pair is assigned to a large margin, and a similar negative pair is assigned to a small margin.

\subsection{Other Backbone Models with CMGSD}
In this section, we answer Q4: To verify the commonality of our proposed method, we performed additional experiments where we chose two other models, \textit{Dual Encoding}\cite{dong2019dual} and \textit{Collaborative Experts (CE) \cite{liu2019use}}, as our backbone models and applied our method to them. We kept all of the settings of backbone models unchanged and used official open-source codes to report the performance of our method.

Table \ref{Other method} shows retrieval results on MSRVTT testing data from \cite{xu2016msr}. Similar to the previous experiments, we observed a substantial increase in all evaluation metrics. Specifically, our method improved the dual encoding by 5.7 on Rsum and gained a 9.6\% relative improvement for CE on Rsum, respectively. The experimental results demonstrated that the proposed CMGSD, which provides enhanced supervision from the video retrieval model itself, is useful and versatile.

\section{Conclusions}
In this paper, we proposed a sample-specific adaptive margin to provide enhanced supervision for video-text representation learning. First, we defined the supervision expert and the rescale function to define the calculation framework for the adaptive margin. Then, we proposed the cross-modal generalized self-distillation method, which is a novel implementation of that framework. Extensive experiments on three benchmarks, MSRVTT, ActivityNet, and LSMDC demonstrated that our proposed method outperformed state-of-the-art methods by a large margin. Moreover, we conducted several experiments using different methods, which indicates that our proposed method can be used as a simple and efficient tool for video retrieval models. In future works, we will explore more information about the model itself and make additional theoretical efforts on our method.

\bibliographystyle{ACM-Reference-Format}
\bibliography{sample-base}


\end{document}